# Evaluating the Efficacy of Large Language Models in Identifying Phishing Attempts

Het Patel
Department of Computer Science
Western University
London, Ontario
hpate384@uwo.ca

Umair Rehman
Department of Computer Science
Western University
London, Ontario
urehman6@uwo.ca

Farkhund Iqbal
College of Technological Innovation
Zayed University
Dubai, United Arab Emirates
farkhund.iqbal@zu.ac.ae

*Abstract*—Phishing, a prevalent cybercrime tactic for decades, remains a significant threat in today's digital world. By leveraging clever social engineering elements and modern technology, cybercrime targets many individuals, businesses, and organizations to exploit trust and security. These cyber-attackers are often disguised in many trustworthy forms to appear as legitimate sources. By cleverly using psychological elements like urgency, fear, social proof, and other manipulative strategies, phishers can lure individuals into revealing sensitive and personalized information. Building on this pervasive issue within modern technology, this paper will aim to analyze the effectiveness of 15 Large Language Models (LLMs) in detecting phishing attempts, specifically focusing on a randomized set of "419 Scam" emails. The objective is to determine which LLMs can accurately detect phishing emails by analyzing a text file containing email metadata based on predefined criteria. The experiment concluded that the following models, ChatGPT 3.5, GPT-3.5-Turbo-Instruct, and ChatGPT, were the most effective in detecting phishing emails.

*Index Terms*—Phishing Email Detection, Large Language Models (LLMs), General Pretrained Transformer (GPT), Bidirectional Encoder Representations from Transformers (BERT), Natural Processing Language (NPL), Social Engineering

## I. INTRODUCTION

In an era where digital communication has become an essential part of daily life, phishing emails have become a prevalent and subtle threat to online security. These fraudulent messages are cleverly designed to be deceptive and manipulative by mimicking legitimate communications [1]. The most used communication channels for introducing such attacks are emails, messaging services, social media platforms, and phone calls [2]. Through these various strategies, phishing attackers can often skillfully lure users into disclosing sensitive information, such as financial information, personal identification details, login credentials, etc. [2], leading to significant privacy and security breaches. To understand the pervasive threat that phishing emails pose today, it is crucial to delve into the historical context and technological advancements that shaped their development. The evolution of phishing can be traced back to the 1990s, a time when digital literacy and the internet were not as modernized and widespread as today [3]. Initially, phishing emails were relatively straightforward, as they relied primarily on the element of surprise and lack of familiarity with online fraud. During this period, phishing primarily consisted of plain text emails, login credentials theft, and trust exploitation [3]. As the internet became more integral and digital literacy improved, phishing tactics were further technologically enhanced. By the mid-2000s, technologies, including HTML and CSS, were utilized to develop content that closely mimicked legitimate communications skillfully [4]. This era also introduced the rise and rapid growth of social media platforms and conglomerates, such as Facebook and Microsoft [5]. With such rapid integration of social media into daily life, these networking platforms quickly became crucial networking and sharing tools. However, extensive involvement and sharing are what introduced significant threats and cybersecurity breaches as malicious content could be shared across networks of followers and facilitated easy accessibility of personal data [6].

While the internet was becoming the central hub for everyday activities, the security measures were often insufficient to handle any malware detection as they lacked the capabilities to identify and counteract any threats. Notably, essential security measures, including 2FA (Two-Factor Authentication), Secure Socket layers (SSL), Sender Policy Framework (SPF), and Domain keys identified mail (DKIM), were not as developed or implemented in many technologies [7]. With these limitations, phishing attacks were more persistent in society, as they lacked multi-layered authentication, unencrypted data transmission, and email sender verification. However, despite the evolution of these technologies, phishing attacks still remain a widespread problem in today's age due to the lack of public awareness and advancements in artificial intelligence [8].

Over the last few years, the capabilities of natural language process models (NLPs) have significantly improved and shaped the ways phishing emails can be created and detected. On one hand, AI-driven techniques and ML algorithms have allowed phishers to generate real and pervasive phishing emails [9]. Models such as GPT-4 and Perplexity AI have effectively demonstrated the ability to perform linguistic tasks at an advanced level and create content that accurately mimics human communication and is indistinguishable by the human eye [10], raising significant concerns about possible misuse and exploitation. The size and parameters that define these models and their complexity, effectiveness, and application areas are growing exponentially [10]. As these parameters expand, the models are growing similarly regarding their processing capability and pattern recognition. Notably, most small language models (SLM) and large language models (LLM) have parameters ranging from 10 billion to over 100 billion [11], impacting both the creation and detection of phishing tactics. While these models, such as GPT-4, Claude AI, and similar LLMs, are initially trained on general datasets, their capabilities can be fine-tuned on specialized data [12], allowing them to adapt to specific needs. This fine-tuning includes adjusting the models to work with domain-specific data sets, creating or retraining Generative Pre-trained Transformers (GPTs) to meet certain requirements, and integrating plugins/API keys with services to improve functionality and automation [13]. From these extensive capabilities, phishers can further exploit these finely tuned models to create deceptive content in several ways. Firstly, by having access to models tailored to certain data sets, such as activity logs, personal information, and financial records, phishers can create more credible and personalized phishing emails [14]. Secondly, with the capability of having a GPT to be modified or created [14], phishers can skillfully develop models specifically designed to mimic the tone, style, format, and other characteristics that legitimate communications use [15]. Lastly, the misuse of API keys, which are tokens that allow for the usage of advanced language process features with services, can allow phishers to use these features in their systems and software [16]. Simultaneously, while AI has enhanced phishing techniques, this new technology has also significantly increased the ability to combat phishing [17]. AI systems are trained to recognize many patterns within large datasets. Besides the tone, style, and format, these systems can interpret many elements, including the metadata in emails, anomaly detection, language detection, and behavioral patterns [18]. By leveraging such data, this paper aims to evaluate the effectiveness and success rate of 15 distinct AI language models in detecting phishing emails.

This study introduces an experiment that started with obtaining a comprehensive text file from Kaggle's fraudulent email database, which includes diverse email metadata [19]. This dataset shortlisted 15 randomized emails and served as the foundation for the analysis. Moving forward, a prompt was developed to guide and standardize the evaluation process across all of the AI models to ensure consistency based on various characteristics. The structured framework of the prompt included factors such as the authenticity of the sender's email address, the urgency and tone of the message, requests for personal or financial information, and many more. Following the prompt development, 15 different AI language models were randomly shortlisted based on two categories: transformer-based models, which received the top priority, and BART-based models. Upon shortlisting these AI models, each model was tested using the subset of data. Every email was entered sequentially for each model to see if the model could accurately identify it as a phishing email with some concrete "yes" or "no" answer. This process involved individually copying and pasting each email into the AI model's interfaces and recording the model's response to see if it correctly identified the email as a phishing attempt.

## II. METHODOLOGY

### A. Data Acquisition

To start the evaluation of language models for identifying phishing emails, the first phase involved acquiring a database of phishing emails to be used as the dataset for analysis. This study used a publicly available dataset from "Kaggle's fraudulent email corpus". The dataset consisted of a 6 MB text file encompassing 4075 fraudulent emails, which contained over 2500 entries from the years 1998–2007. This subset of entries pertained to emails related to the 419 scam, also known as the Nigerian fraud letters. Following this acquisition, a data cleaning process was done to ensure that the data was reliable, diverse, and representative of the various phishing strategies. For this study, a subset of 50 emails were randomly selected and verified from the text file. However, considering that many AI models have processing limitations, a smaller subset of 15 emails was randomly sampled from the verified group for the analysis to ensure efficient management and optimization. The email entries included many phishing tactics, metadata, and other characteristics. Specifically, the dataset contained the following information: sender information, recipient details, subject lines, and message content. The dataset was obtained from Kaggle's Fraudulent Email Corpus [19], available at https://www.kaggle.com/datasets/rtatman/fraudulent-email-corpus.

### B. Development of Evaluation Prompt

Following the data acquisition, a prompt was developed to standardize the evaluation process across all the AI models to ensure consistency based on various characteristics. This structured framework was used to systematically assess each email against key indicators that are commonly related to phishing. Such indicators include: the authenticity of the sender's email address, the urgency and tone of the message, the accuracy of sentence structure and grammar, requests for personal or financial information, narrative consistency, and plausibility of any offers made, among many more. The specific prompt that was provided for evaluating the emails is as follows: "Given an email with the following characteristics: sender's email address authenticity, urgency and tone, spelling and grammar accuracy, requests for personal or financial information, unsolicited nature of the contact, plausibility of offers, personalization of greetings, presence of detailed contact information, and narrative consistency, evaluate and summarize the likelihood of the email being a phishing attempt. Consider the presence of red flags such as implausible offers, requests for sensitive information, and generic or unsolicited contact, and provide a brief summary judgment on the email's legitimacy." Furthermore, alongside the evaluation prompt, a numerical scale ranging from 1-10 was provided to quantitatively assess the likelihood of each email being a phishing attempt. A rating of '1' on this scale signified a very low likelihood of the email being a phishing attempt, indicating that the email is legitimate and does not have any of the common characteristics related to phishing. However, a rating of "10" signified a very high likelihood of the email being a phishing attempt, indicating that the email does contain multiple characteristics related to phishing. The specific prompt provided for evaluating the phishing attempts is as follows: "Then, provide a numerical summary of the email's legitimacy on a scale from 1 to 10, where 1 signifies a very low likelihood of being a phishing attempt (the email is entirely legitimate), and 10 signifies a very high likelihood of being a phishing attempt (the email is almost certainly fraudulent)."

### C. Selection of Language Models:

Following the development of the evaluation prompt, the AI models' selection process was designed to focus on diversity and representativeness. Specifically, 15 AI models were shortlisted from Poe (https://poe.com/), an AI chat platform that allows interactions with various AI models. The selection process involved a systematic approach, where the AI models were chosen based on their size and processing capabilities, ranging from high to low. The largest and most comprehensive models were prioritized first in the selection process, ensuring that the large language models (LLMs) received the most representation due to their increased analytical capabilities and parameter size. Subsequently, the selection process was further narrowed down to models of intermediate and smaller sizes to ensure a balanced representation. This was done to provide an overview that includes models with high processing capabilities and those with lower processing capabilities. The goal was to analyze the effectiveness of the AI models in identifying phishing attempts across varying computational power and abilities. The following models were chosen: ChatGPT 3.5, GPT-3.5-Turbo-Instruct, ChatGPT, Code-Llama-7b, ChatGPT 4, Qwen-72b-chat, Assistant, Solar-Mini, Claude-2-100k, Llama-2-70b-Groq, Google-paLM, Claude-Instant, Code-Llama-34b, Mistral-Medium, fw-mistral-7b.

### D. Evaluation Process:

The evaluation process was then conducted to analyze the effectiveness of the selected AI models in identifying phishing attempts. This process involved systematically parsing each of the 15 emails through the interfaces of the 15 AI models. The goal was to observe and record the models' abilities to correctly classify each email as either phishing or legitimate based on the given criteria and the prompt. For every model evaluated, each email was analyzed and followed up by a detailed assessment summary, along with a numerical score ranging from 1 to 10 indicating the likelihood of a phishing attempt. Additionally, to clearly present the results, multiple histograms and a correlation matrix were used to illustrate the relationships between the phishing confidence scores assigned by various AI models.

## III. RESULTS

Table I presents the performance of various language models in identifying phishing emails from a dataset composed entirely of such emails. Each model was assessed on its ability to analyze email content and metadata to detect phishing attempts. Higher scores indicate better performance in identifying these malicious emails. The variation in scores across models highlights differences in their detection capabilities, with ChatGPT 4 showing the highest average effectiveness in this specific task. Factors such as the datasets on which they were trained, their underlying algorithms, model architectures, and patterns all influenced the variation.

In the context of this study, confidence scores refer to the numerical values that the LLMs assign to each email, representing the models' assessments of the likelihood that an email is a phishing attempt. These scores range on a predefined scale from 1 to 10, where a higher score indicates a greater probability that the email is a phishing attack. The confidence score is a direct outcome of the model's analysis, reflecting how closely the content and characteristics of an email align with the model's learned indicators of phishing. Additionally, the "density" here refers to how these confidence scores are distributed across the evaluated emails, as seen through the histograms' Kernel Density Estimation (KDE) curves in Figure 1. For instance, a higher density in the higher score range would imply that many emails were assessed with high confidence scores, suggesting phishing. Conversely, a higher density in the lower score range would imply that many emails were assessed with low confidence scores, indicating they are less likely to be considered phishing attempts. A general tendency towards higher scores was noted among most models, reflecting a strong confidence level in accurately identifying phishing emails. However, despite this general inclination towards higher scores, some models displayed a broader distribution across the scoring range, suggesting a more refined approach to the assessment. The analysis indicated the following three models as the best performing in phishing detection: ChatGPT 3.5, GPT-3.5-Turbo-Instruct, and ChatGPT. These Transformer models consistently scored higher on the confidence score scale, with most evaluated emails receiving scores between 8 and 10. However, the five least effective models, Mistral Medium, fw-mistral-7b, Llama-2-70b-Groq, Claude-2-100k, and Claude-Instant, showed a distribution of confidence scores for phishing in the lower range, with most emails receiving scores between 2 and 10. The remaining models, including ChatGPT, Assistant, Qwen-72b-chat, Code-Llama-7b, Code-Llama-34b, Google-paLM, and Solar-Mini, were categorized within the intermediate range of effectiveness for detecting phishing attempts, with confidence scores ranging from 6 to 10. This variability and divergence in scoring patterns highlight the need for model calibration. Such tuning is crucial because it enables a balanced and effective phishing detection strategy, ensuring higher sensitivity to threats and minimizing false positives.

Additionally, a correlation matrix was used, as shown in

TABLE I
SUMMARY OF LANGUAGE MODELS' PERFORMANCE IN PHISHING EMAIL DETECTION

| AI Model | Count | Mean | STD | Min | 25% | Median | 75% | Max |
|---|---|---|---|---|---|---|---|---|
| GPT-3.5 Score | 15 | 9.00 | 0.00 | 9 | 9 | 9 | 9 | 9 |
| GPT-3.5-Turbo-Instruct Score | 15 | 9.00 | 0.00 | 9 | 9 | 9 | 9 | 9 |
| ChatGPT Score | 15 | 9.07 | 0.59 | 8 | 9 | 9 | 9 | 10 |
| ChatGPT 4 Score | 15 | 9.87 | 0.52 | 8 | 10 | 10 | 10 | 10 |
| Assistant Score | 15 | 8.80 | 0.41 | 8 | 9 | 9 | 9 | 9 |
| Qwen-72b-chat Score | 15 | 8.93 | 0.26 | 8 | 9 | 9 | 9 | 9 |
| Claude-2-100k Score | 15 | 8.27 | 2.15 | 1 | 8 | 9 | 9 | 10 |
| Claude-Instant Score | 15 | 8.77 | 2.29 | 1 | 8.5 | 9.5 | 10 | 10 |
| Code-Llama-7b Score | 15 | 7.73 | 1.33 | 6 | 7 | 7 | 8 | 10 |
| Code-Llama-34b Score | 15 | 8.40 | 0.83 | 7 | 8 | 8 | 9 | 10 |
| Llama-2-70b-Groq Score | 15 | 7.67 | 2.55 | 1 | 8 | 9 | 9 | 9 |
| Google-paLM Score | 15 | 8.20 | 0.41 | 8 | 8 | 8 | 8 | 9 |
| Solar-Mini Score | 15 | 8.33 | 0.72 | 7 | 8 | 8 | 9 | 9 |
| Mistral-Medium Score | 15 | 9.07 | 2.28 | 1 | 9 | 10 | 10 | 10 |
| fw-mistral-7b Score | 15 | 8.00 | 2.27 | 2 | 7 | 9 | 9 | 10 |

Figure 2, to illustrate the relationships between the phishing confidence scores assigned by the various AI models. From this correlation matrix, each cell displayed the correlation coefficient between the two models' scores, with values ranging from -1 to 1. A value close to 1 indicated a strong positive correlation, suggesting that as the score from one model increases, the score from another model also tends to increase. Conversely, a value close to -1 indicated a strong negative correlation, and a value around 0 suggested little to no linear relationship between the models' scores. High positive correlations were observed among many model pairs, with coefficients significantly above 0.5, indicating that these models tend to agree on their phishing email assessments. This suggested that models with high correlations utilized similar criteria or had similar sensitivities in detecting phishing threats. The moderate to strong positive correlations across the matrix highlighted a consensus among the AI models on the phishing likelihood of emails, which is crucial for a reliable phishing detection framework. Despite most positive correlations, their strengths varied, highlighting differences in their underlying architectures and parameters.

## IV. DISCUSSION

Among the 15 evaluated LLMs, the top performers in detecting phishing attempts were ChatGPT, GPT-3.5-Turbo-Instruct, and ChatGPT 4. These models utilized a decoder-only architecture, GPT [20]. Conversely, the five models that demonstrated the lowest effectiveness, each utilizing distinct architectures, were Mistral Medium, fw-mistral-7b, Llama-2-70b-Groq, Claude-2-100k, and Claude-Instant. Notably, Claude-2-100k and Claude-Instant were built on the BERT architecture, which is encoder only [21], in contrast to Mistral Medium, fw-mistral-7b, and Llama-2-70b-Groq, which integrated both encoder and decoder elements into their frameworks [22]. There are three critical differences between BERT and GPT families of models: their training objectives, the computational scale at which they are trained, and their prompt size. At its core, the GPT architecture primarily focuses on generating text by predicting the upcoming words in a sequence given all the previous words [23]. By emphasizing this prediction mechanism, models with the decoder-only architecture demonstrated a profound strength in understanding and generating language patterns during the study. This predictive capability allowed them to effectively grasp the flow of natural language [24], making them ideal for identifying any inconsistencies indicative of phishing attempts and foreseeing potential fraudulent scenarios. However, the models based on the BERT architecture, prioritizing encoding over decoding, performed less adeptly than anticipated in tasks like phishing detection [25], where generating text and predicting continuations are essential. Unlike the GPT architecture, BERT analyzes text using a masked language modeling system in a bidirectional manner [26], meaning it simultaneously considers both the preceding and succeeding context of a word. This bidirectional nature of BERT models makes them powerful for tasks that require a thorough understanding of the text, such as sentiment analysis [26], where the sentiment of a word can depend heavily on the words around it, or question-answering systems, where the answer needs to consider the context that is provided in both the question and the surrounding information. However, to achieve peak performance on the tasks, BERT models often require task-specific fine-tuning [27]. This process involves additional training where the model is adjusted and optimized using a smaller, task-specific dataset after its initial pre-training [28]. This fine-tuning step is crucial for BERT models to effectively apply their bidirectional understanding of the text to a given task's specific nuances and requirements. Additionally, the scalability of these LLMs and the data on which they were trained directly correlate with their effectiveness [29]. The parameters in a language model are the foundational elements that enable it to recognize complex patterns and more extensive dependencies [30].

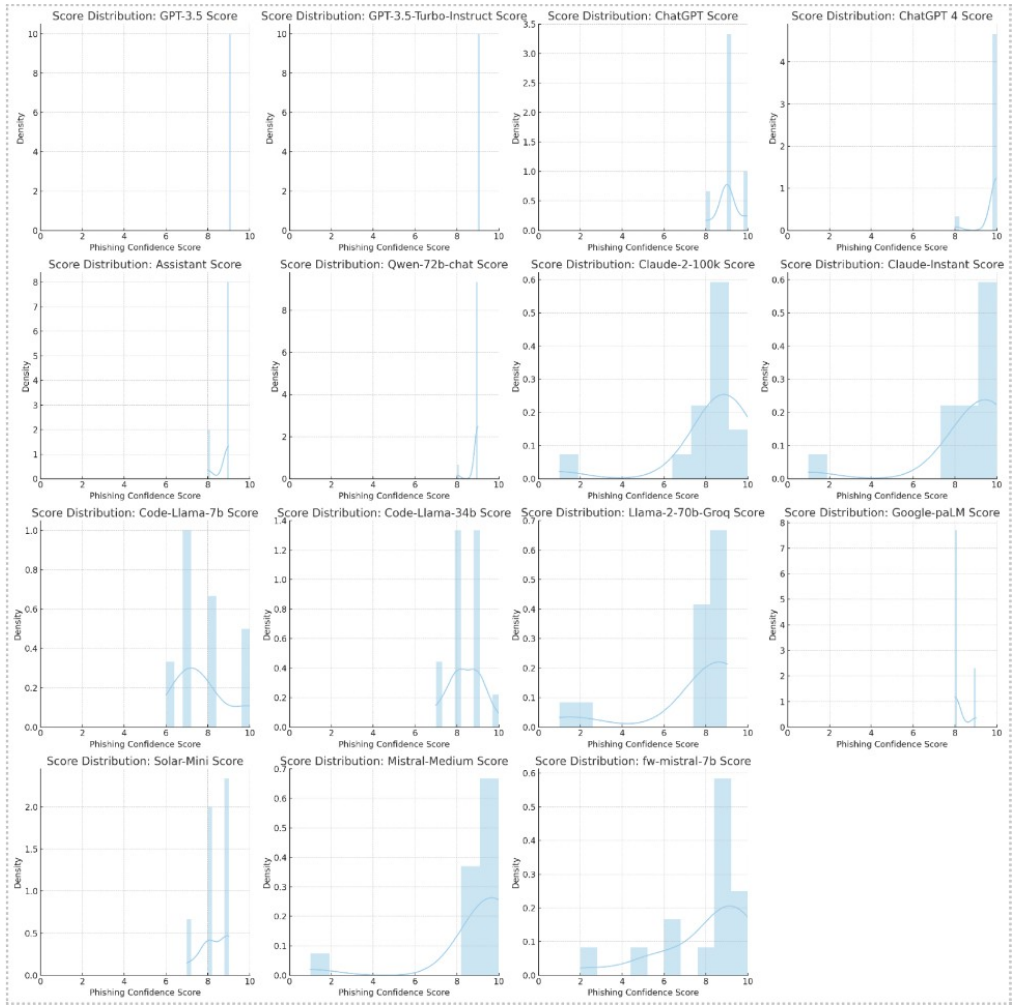

Fig. 1. The Effectiveness of Various AI Models in Identifying Phishing Emails

In terms of the model and parameter size among the leading models, both ChatGPT 3.5 and ChatGPT 3.5-Turbo-Instruct were equipped with over 175 billion parameters [31]. At the same time, ChatGPT 4 significantly exceeded this, having more than 1.7 trillion parameters [30]. However, among the five least effective models, parameter sizes varied, with Mistral Medium at the lowest with 7 billion parameters [32] and the Anthropic models at the highest with over 130 billion [33]. This significant variance in parameter size between the top-performing models and the least effective ones highlights the importance of computational capacity for specialized tasks such as phishing detection. Furthermore, another critical advantage of GPT models over BERT models is their capability to perform Zero-/few-shot learning [34]. Unlike the BERT models, which require fine-tuning for specific tasks, this feature allows the GPT models to execute tasks without explicit fine-tuning and little to no extensive task-specific training [35]. The GPT series can, therefore, directly apply their generalized pre-trained knowledge to identify phishing attempts. This means that these GPT models can effectively analyze and assess the legitimacy of communications to identify any potential phishing threats based on their understanding of language patterns, context, and anomalies learned during the training phase [36].

## V. CONCLUSION

This paper explores the effectiveness of 15 Large Language Models (LLMs) in detecting phishing attempts, focusing on 15 scam emails chosen randomly from a dataset of 419 scam emails. The predictive nature of these models, which enables them to detect potential fraudulent scenarios based on an input prompt, proved particularly effective against the many manipulative tactics in fraudulent communications. The evaluation revealed that OpenAI-developed models, such as ChatGPT, GPT-3.5-Turbo-Instruct, and ChatGPT, excel in identifying phishing emails. Their decoder-only architecture and extensive parameter sizes allow them to understand nuanced language patterns and context better. This understanding is crucial for phishing detection, where distinguishing between legitimate and fraudulent communications can be incredibly subtle. Leveraging the predictive nature and contextual awareness of these models can make phishing detection systems more accurate and efficient. In contrast, models like Mistral Medium, fw-mistral-7b, Llama-2-70b-Groq, Claude-2-100k, and Claude-Instant, which use different architectures or have less computational capacity, performed less efficiently. These insights highlight a significant opportunity to refine and improve the cybersecurity measures in use. Harnessing the predictive and contextual capabilities of models like the ChatGPT series can significantly boost the accuracy and efficiency of phishing detection systems. By leveraging the deep learning ability and the NLP strengths these models use, system training can be improved to recognize and flag potential phishing threats with greater precision, even when the phishing attempts use sophisticated language or contextual clues that traditional systems may not easily detect. Building on these advancements, future studies could try to further refine and expand the capabilities of these models in a cautionary manner. One way to achieve this is by creating self-learning algorithms that adapt to new threats in real time, without requiring extensive retraining.

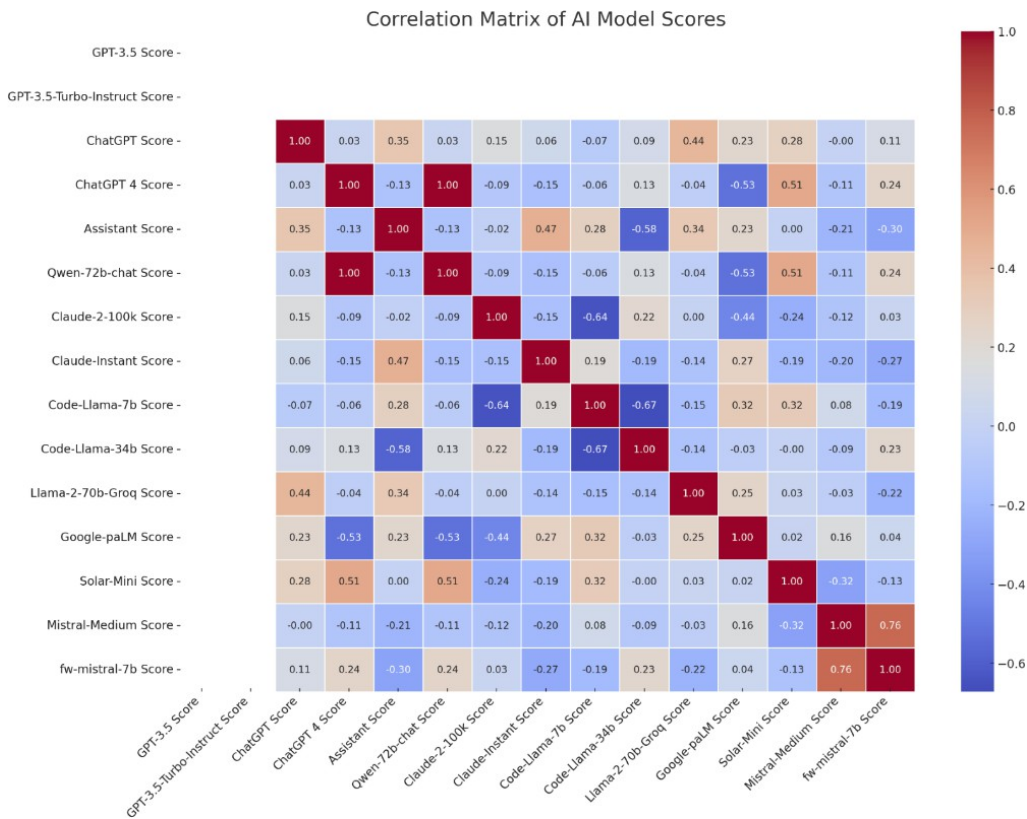

Fig. 2. An illustration depicting the correlations among the phishing confidence scores assigned by different AI models